\documentclass[10pt,journal,
compsoc]{IEEEtran}

\usepackage{color}
\usepackage{graphicx}
\usepackage{amsmath,amssymb}
\usepackage{times}
\usepackage{multirow}
\usepackage[caption=false]{subfig}
\usepackage{ragged2e}
\usepackage{multirow,array}
\usepackage{cite}
\usepackage{rotating}
\usepackage{multicol,multirow}
\usepackage{overpic}
\usepackage{contour}
\usepackage{booktabs}       

\usepackage{xcolor}
\usepackage{epsfig}
\usepackage{graphicx}
\usepackage{soul}

\usepackage[colorlinks,bookmarksnumbered,bookmarksopen,citecolor=blue]{hyperref}
\usepackage{silence}
\usepackage{pifont}
\def\eg{\emph{e.g.}}
\def\ie{\emph{i.e.}}
\def\etal{et al.}

\def\wrt{\emph{w.r.t.~}}

\renewcommand{\paragraph}[1]{\vspace{3mm} \noindent \textbf{#1}}

\newcommand{\figref}[1]{Fig.~\ref{#1}}
\newcommand{\tabref}[1]{Tab.~\ref{#1}}
\newcommand{\equref}[1]{Eqn.~\ref{#1}}
\newcommand{\secref}[1]{Sec.~\ref{#1}}

\begin{document}

\title{SC-DepthV3: Robust Self-supervised Monocular Depth Estimation for Dynamic Scenes}

\author{Libo Sun$^*$, Jia-Wang Bian$^*$, Huangying Zhan, Wei Yin, Ian Reid, Chunhua Shen\\[0.2cm]
\IEEEcompsocitemizethanks{
  \IEEEcompsocthanksitem 
  First two authors contributed equally.
  J.-W. Bian is the corresponding author. He is with the University of Oxford, United Kingdom;
  \IEEEcompsocthanksitem L. Sun and I. Reid are with The University of Adelaide, Australia;
  \IEEEcompsocthanksitem H. Zhan is with the OPPO US Research Center, United States.
  \IEEEcompsocthanksitem W. Yin is with the DJI Technology, China;
  \IEEEcompsocthanksitem C. Shen is with Zhejiang University, China. 
  \IEEEcompsocthanksitem Part of this work was done when J.-W. Bian, H. Zhan, W. Yin, and C. Shen were with the University of Adelaide;
}
}

\markboth{IEEE Transactions on Pattern Analysis and Machine Intelligence}%
{Sun and Bian \MakeLowercase{\textit{et al.}}: Robust Self-supervised Monocular Depth Estimation for Dynamic Scenes}

\IEEEtitleabstractindextext{%
\begin{abstract}
\justifying
Self-supervised monocular depth estimation has shown impressive results in static scenes.
It relies on the multi-view consistency assumption for training networks,
however, that is violated in dynamic object regions and occlusions.
Consequently, existing methods show poor accuracy in dynamic scenes,
and the estimated depth map is blurred at object boundaries because they are usually occluded in other training views.
In this paper, we propose SC-DepthV3 for addressing the challenges.
Specifically,
we introduce an external pretrained monocular depth estimation model for generating single-image depth prior,
namely pseudo-depth,
based on which we propose novel losses to boost self-supervised training.
As a result, our model can predict sharp and accurate depth maps, 
even when training from monocular videos of highly dynamic scenes.
We demonstrate the significantly superior performance of our method over previous methods on six challenging datasets,
and we provide detailed ablation studies for the proposed terms.
Source code and data have been released at \url{%
https://github.com/JiawangBian/sc_depth_pl}
\end{abstract}

}
\maketitle

\begin{figure*}[t]
  \centering
  \includegraphics[width=1.0\linewidth]{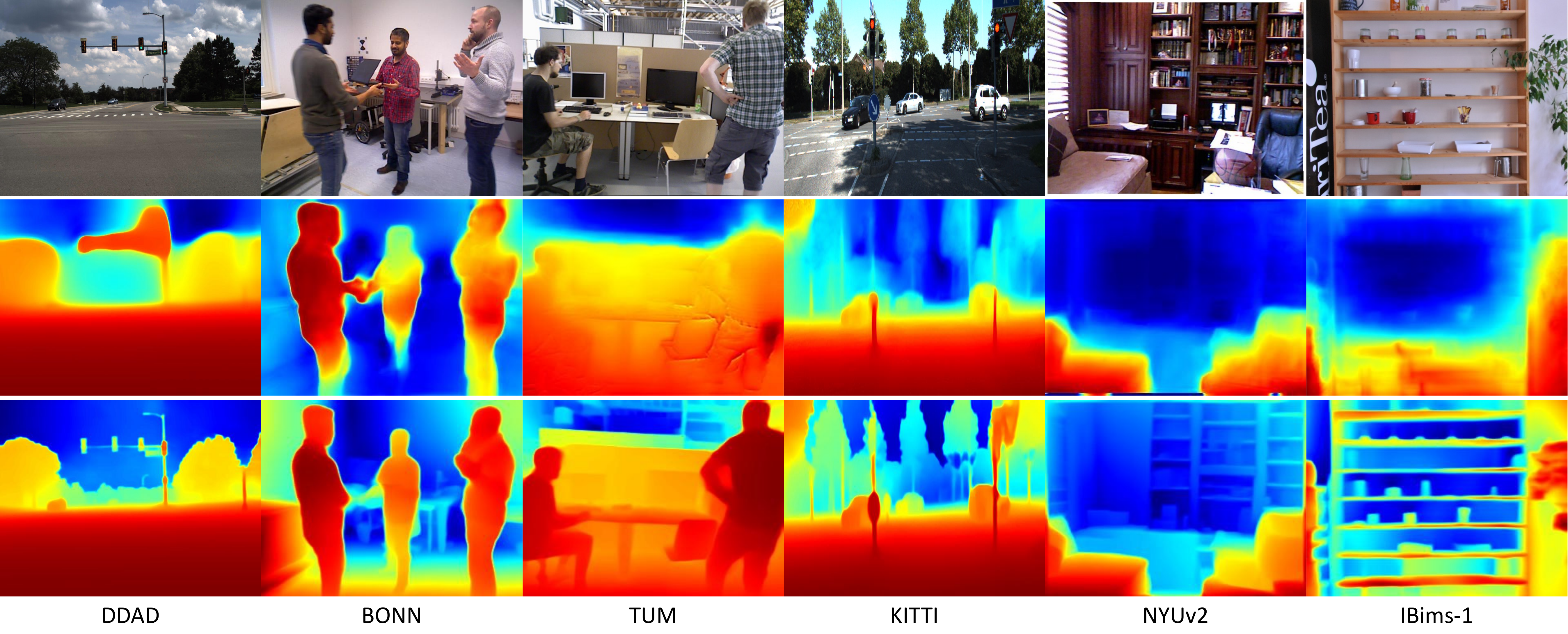}
  \caption{Qualitative monocular depth estimation results on six datasets.
  We compare our method (third row) with SC-Depth~\cite{bian2021ijcv} (second row),
  which is one of the previous state-of-the-art self-supervised methods.
  Compared with it, our method enables more robust learning in dynamic scenes (left three columns) and generates sharper depth maps, 
  particularly at object boundary areas.
  }\label{fig:ours-vis}
\end{figure*}

\IEEEraisesectionheading{\section{Introduction}\label{sec:intro}}
\IEEEPARstart{M}{onocular} depth estimation~\cite{eigen2014depth} has attracted great attention in computer vision.
It provides valuable cues for various downstream tasks,
such as semantic image segmentation~\cite{park2017rdfnet}, salient object detection~\cite{zhou2020rgbd}, 3D reconstruction~\cite{newcombe2011kinectfusion}, novel view synthesis~\cite{bian2021nvss},
and visual odometry~\cite{zhan2019dfvo,bian2021ijcv}.
Early work~\cite{eigen2014depth, liu2016learning} solves the monocular depth estimation problem by using supervised learning. 
However, these methods rely on ground-truth depth labels that are not always available in real-world scenes. 
To address this limitation, self-supervised monocular depth estimation methods were proposed and showed
that a depth network could be trained from stereo image pairs~\cite{garg2016unsupervised} or 
monocular videos with ego-motions~\cite{zhou2017unsupervised}
without the need for ground-truth depth labels.
We focus on self-supervised learning of monocular depth from videos since only a single camera is required to collect training data in this setup,
which has great potential for advancing real-world applications.

Self-supervised methods typically rely on the multi-view consistency assumption for training networks,
\eg, the photometric loss~\cite{zhou2017unsupervised} and geometry consistency loss~\cite{bian2021ijcv} that were used in previous methods.
This assumption provides effective constraints for learning scene geometry,
while it is violated at regions with occlusion (\eg, object boundaries) and moving objects.
Therefore, existing methods often show only excellent results in (almost) static scenes such as KITTI~\cite{Geiger2013IJRR} and NYUv2~\cite{silberman2012indoor} datasets.
When training on more challenging dynamic datasets that have an amount of fast-moving objects,
previous state-of-the-art methods~\cite{monodepth2,packnet,bian2021ijcv} show poor accuracy.
Moreover, the estimated depth map is blurred at object boundaries because they are usually occluded in other training views.
We illustrate several examples of qualitative monocular depth results in~\figref{fig:ours-vis}.

To address the issues caused by moving objects and occlusions,
existing approaches usually detect these bad regions and then exclude them from training.
The methods can be categorized into four classes according to how they detect dynamic regions,
involving in the prediction-based~\cite{zhou2017unsupervised},
semantic-based~\cite{casser2019struct2depth, gordon2019depth, packnet-semguided,klingner2020self}, 
flow-based~\cite{yin2018geonet,zou2018df, ranjan2019cc, chen2019self}, 
and geometry-based~\cite{bian2021ijcv}.
These methods can reduce corruption from noisy losses during training and generally improves overall accuracy, 
however, it leads to poor results on dynamic regions at inference time because these regions are not sufficiently regularized in training.
There are also more sophisticated approaches \cite{lee2021learning, li2020unsupervised} that model the velocity of each moving object in multiple views, but they rely on solving a challenging problem in themselves.

We propose SC-DepthV3 in this paper,
which addresses the above-mentioned issues by leveraging external single-image constraints.
Specifically, we leverage an off-the-shelf monocular depth estimation model~\cite{yin2021learning} to generate the single-image depth prior,
which we 
term 
\emph{pseudo-depth}.
Based on it, we propose effective losses to constrain the depth estimation network in self-supervised learning.
Here, we use LeReS~\cite{yin2021learning} for generating pseudo-depth,
which is trained in large-scale datasets with supervised learning and enables zero-shot generalization in previously unseen data.
The excellent qualitative results have been demonstrated in \cite{yin2021learning},
while we find that pseudo-depth may show low quantitative accuracy.
\figref{fig:pseudo-depth-vis} gives an example, where we visualize the error map of pseudo-depth by comparing it with the ground truth.
This phenomenon makes supervised zero-shot methods unsuitable for accuracy-sensitive tasks such as visual SLAM and 3D Reconstruction.
Furthermore, as pseudo-depth is not quantitatively accurate,
it is non-trivial to use it for boosting self-supervised learning.
In this paper, our technical contribution is designing effective losses that use imperfect pseudo-depth.
It is also worth mentioning that although we use external depth estimation networks, 
they are only trained once and can be used as off-the-shelf tools in new scenes.
Therefore, in practice, our method does not add extra cost to purely self-supervised methods.

The key to solving the dynamic region issue is the proposed \textbf{Dynamic Region Refinement} (DRR) module.
The method is inspired by an observation, \ie, we find that pseudo-depth maintains excellent depth ordinal (the further/nearer relations) between any two objects or pixels.
To capitalize on these findings,
we propose to extract the ``ground-truth" depth ordinal information between dynamic and static regions (from pseudo-depth) and use it to regularize the self-supervised depth estimation in dynamic regions.
Specifically, we sample point pairs between two regions and apply depth ranking loss~\cite{chen2016single}.
This is effective because the static backgrounds have already been well-supervised by multi-view losses,
and the dynamic regions could be uniquely localized by sampling sufficient point pairs between dynamic and static regions.
Our method is also based on the fact that the depth ordinal in pseudo-depth is sufficiently accurate~\cite{yin2021learning}.
Furthermore, to segment dynamic regions from static backgrounds,
we use the self-discovered mask that was proposed in SC-Depth~\cite{bian2021ijcv} and generated by computing forward-backward depth inconsistency in self-supervised training,
so the external segmentation networks are not required.
\figref{fig:drr-vis} illustrates the proposed DRR module.

Moreover, we observe that pseudo-depth shows smooth local structures and clean object boundaries.
This motivates us to propose a \textbf{Local Structure Refinement} (LSR) module to improve the self-supervised depth estimation \wrt depth details.
The proposed module contains two parts.
On the one hand, we extract the surface normal from both pseudo-depth and network-predicted depth,
and we constrain them to be consistent by applying a normal matching loss.
This improves the overall depth significantly.
On the other hand, we constrain depth estimation at object boundary areas by applying our proposed relative normal angle loss.
More specifically, we sample point pairs around image edges and enforce their relative normal angles to be consistent between pseudo-depth and self-supervised depth.
As a result, our method improves qualitative depth estimation results significantly,
particularly at object boundaries.
\figref{fig:ours-vis} shows several examples of the qualitative depth estimation results.

Our contributions are as follows:
\begin{itemize}
    \item We propose SC-DepthV3 for robust self-supervised learning of monocular depth in highly dynamic scenes, which allows for predicting accurate and sharp depth maps.
    \item We propose Dynamic Region Refinement (DRR) and Local Structure Refinement (LSR) modules, which are based on pseudo-depth to boost self-supervised learning.
    \item We conduct comprehensive experiments and ablation studies on six challenging datasets. The results demonstrate the efficacy of our proposed methods.
\end{itemize}

\section{Related Work}
\label{sec:related_work}

\paragraph{Self-supervised Monocular Depth Estimation.}
Garg~\etal \cite{garg2016unsupervised} proposed to train monocular depth estimation models on stereo image pairs by using the photometric loss.
Zhou~\etal~\cite{zhou2017unsupervised} proposed to train the depth estimation model on videos by jointly training a pose estimation model.
Following them, many advanced techniques~\cite{godard2017unsupervised, zhan2018unsupervised,  mahjourian2018unsupervised, yin2018geonet, chen2019self, monodepth2, bian2021ijcv, packnet, packnet-semguided,bian2021tpami} were proposed to boost the performance.
However, multi-view ambiguities make the self-supervised method hard to handle dynamic objects and object boundaries.
Previous methods either excluded these regions from training~\cite{yin2018geonet, casser2019struct2depth, gordon2019depth, bian2021ijcv} or modeled the object motions~\cite{lee2021learning,li2020unsupervised}, but both solutions have their drawbacks.
More specifically, simply excluding dynamic regions would result in poor accuracy on these regions at the inference time,
and modeling each object's motion is ill-posed and may not be robust in dynamic scenes.
Compared with them, our method leverages pretrained single-image prior for resolving multi-view ambiguities,
leading to a SOTA self-supervised depth estimation method.
More recent methods include \cite{wang2023planedepth, si2023fully, bangunharcana2023dualrefine}.

\begin{figure*}[t]
  \centering
  \includegraphics[width=0.99\linewidth]{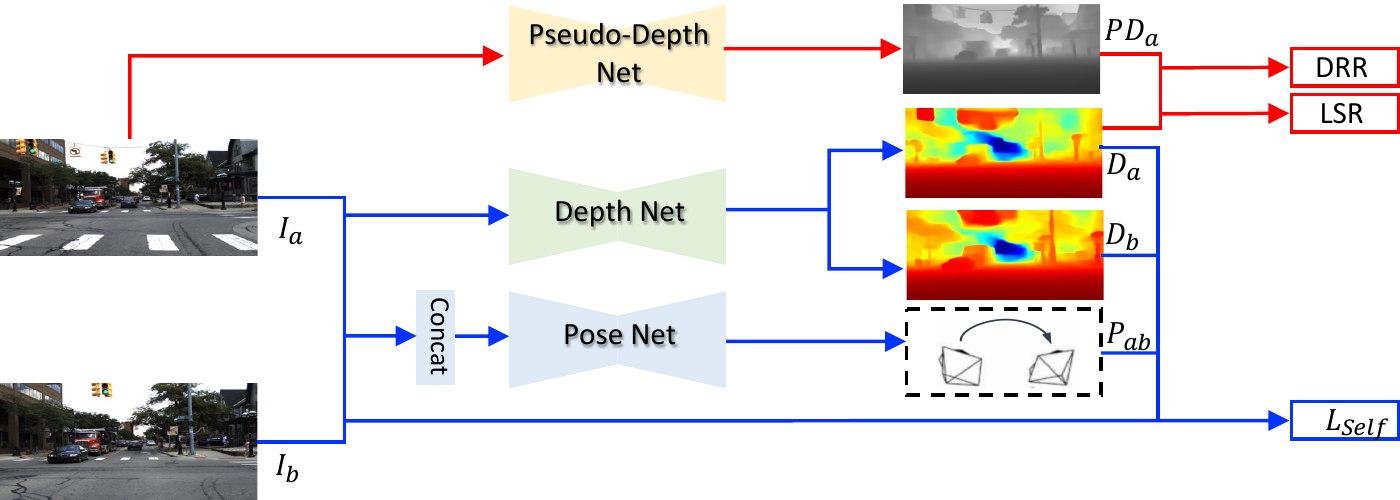}
  \caption{Method overview. Firstly, given a training sample (\ie, $I_a$ and $I_b$ two images), 
  we follow SC-Depth~\cite{bian2021ijcv} to compute self-supervised losses $L_{Self}$ (\equref{eqn-totalloss}), which is described in \secref{sec:baseline}.
  Secondly, we generate pseudo-depth $PD_a$ using a pretrained depth estimation network, which is discussed in \secref{sec:prior}.
  Finally, we propose DRR and LSR modules to constrain the network prediction ($D_a$) by using $PD_a$,
  which are presented in \secref{sec:dynamic} and \secref{sec:structure}, respectively.
  }
  \label{fig:method-vis}
\end{figure*}

\paragraph{SC-Depth Series Methods.}
This paper is the third version of the SC-Depth series methods.
In the SC-Depth~\cite{bian2021ijcv}, we addressed the scale inconsistency issue, 
so our method enables scale-consistent depth estimation over the video,
which is beneficial to video-based tasks such as Visual SLAM.
In the SC-DepthV2~\cite{bian2021tpami}, we analyzed the rotation issue in videos that are captured by handheld cameras,
and we proposed an auto-rectify network to handle the large rotation.
The V1 and V2 have shown great accuracy in both indoor and outdoor scenes.
However, their predicted depth maps are blurred at object boundaries, 
and they suffer in highly dynamic scenes.
In this paper, we propose SC-DepthV3 address the issue of dynamic objects and blurred object boundaries.

\paragraph{Zero-shot Monocular Depth Estimation.}
Many existing methods leverage large-scale datasets and supervised training~\cite{xian2018monocular,li2018megadepth,li2019learning,wang2019web,chen2019learning,yin2020learning,yin2021learning, ranftl2020towards} to train monocular depth estimation models towards zero-shot generalization on unseen data.
For example, \cite{xian2018monocular,li2018megadepth,li2019learning,wang2019web,chen2019learning,yin2020learning} collect stereo images/videos from the internet and use geometric reconstruction tools~\cite{hirschmuller2005accurate,schoenberger2016mvs} to generate dense ground-truth depth labels.
\cite{ranftl2020towards} export perfect ground-truth depths from the synthetic 3D movies~\cite{Butler_ECCV_2012}.
Recently, LeReS~\cite{yin2021learning} and DPT~\cite{Ranftl2021} achieve the state-of-the-art performance.
However, note that their predicted depths are scale-shift-invariant, due to the high diversity of different scenes,
which show low quantitative accuracy in out-of-distribution data and cannot be used for 3D reconstruction.
Nevertheless, we find that their predicted depths carry good attributes that could be leveraged for boosting self-supervised learning of monocular depth estimation.
Compared with these methods, our method enables consistent and accurate depth estimation for video-based tasks such as Visual SLAM, 
which has been demonstrated in SC-Depth~\cite{bian2021ijcv}, thanks to the scale-consistency constraints.

\paragraph{Knowledge Transfer.}
Our method is also related to knowledge transfer approaches,
because the proposed method can be regarded as transferring the knowledge of pretrained monocular depth estimation models~\cite{yin2021learning} to our self-supervised trained models.
However, we argue that our method is very different from previous knowledge transfer or distillation methods.
On the one hand, knowledge is often transferred by finetuning pretrained models in new datasets~\cite{luo2020consistent},
which is not our case and cannot solve the challenges in our problem.
The main issue in our problem is the imperfect self-supervised loss,
so even if we finetune pretrained models (\ie, it provides a good initialization),
the model would become worse and worse with training due to the deficient self-supervised loss functions.
On the other hand, knowledge transfer could also be achieved by conducting semi-supervised learning on mixed datasets.
Specifically, we can train models on both previous large-scale datasets with ground-truth labels and new datasets without annotations,
and then we apply supervised loss in the former and self-supervised loss in the latter.
However, this involves new challenges of mix-data training, long training time, and the maintenance of large-scale previous data.
In contrast, our method is more elegant than semi-supervised training,
since the teacher model is trained only once on large-scale datasets and can be used as an off-the-shelf tool to generate pseudo-depth in new scenes.
Moreover, our student model shows significantly higher accuracy than the teacher model,
which is rare in the field of knowledge distillation.


\section{Method}
\label{sec:method}

\figref{fig:method-vis} illustrates an overview of the proposed method.
First, our method is based on SC-Depth~\cite{bian2021ijcv} for basic self-supervised training,
which we describe in detail in \secref{sec:baseline}.
Second, we discuss the single-image depth prior in \secref{sec:prior} that is generated by using the off-the-shelf monocular depth estimation methods and used in our method for generating auxiliary supervision signals.
Finally, we describe the Dynamic Region Refinement (DRR) in \secref{sec:dynamic} and Local Structure Refinement (LSR) modules in \secref{sec:structure}, respectively,
which are the proposed terms to boost self-supervised training.

\subsection{Self-supervised Depth Learning (SC-Depth)}
\label{sec:baseline}

In the self-supervised learning framework, a monocular depth estimation network (DepthNet) and a relative 6-DoF camera pose estimation network (PoseNet) are jointly trained on a large number of monocular videos.
First, given a consecutive image pair ($I_a$, $I_b$) randomly sampled from a training video,
we predict their depths ($D_a$, $D_b$) by forwarding the DepthNet and estimate their relative 6-DoF camera pose $P_{ab}$ by forwarding the PoseNet.
Then, we generate the warping flow between two images using the predicted depth and pose,
followed by synthesizing the $I'_a$ using the flow and $I_b$ via bi-linear interpolation.
Finally, we penalize the color inconsistencies between $I_a$ and $I'_a$,
and we also constrain the geometry consistency between $D_a$ and $D_b$,
which back-propagates the gradients to the networks.
The objective function is described below.

First, we use the geometry consistency loss $L_G$~\cite{bian2021ijcv} to encourage the predicted depths ($D_a$, $D_b$) to be consistent with each other in 3D space.
Formally, 
\begin{equation}\label{eqn-gc}
L_{G} = \frac{1}{|\mathcal{V}|} \sum_{p \in \mathcal{V}} D_{\text{diff}}(p),
\end{equation}
where $V$ stands for valid points that are projected inside the image.
$D_{\text{diff}}$ stands for the pixel-wise depth inconsistency between $D_a$ and $D_b$,
which is detailed explained in ~\cite{bian2021ijcv}.
With it, we can obtain the self-discovered mask:
\begin{equation}\label{eqn-mask}
M_s = 1 - D_{\text{diff}},
\end{equation}
which assigns lower weights to dynamics and occlusions than static regions, 
since the former is geometrically inconsistent across multiple views.
We use this mask in our proposed DRR module(~\secref{sec:dynamic}) to localize dynamic regions.

Second, we use the weighted photometric loss $L_{P}^M$ to constrain the warping flow between $I_a$ and $I_b$ that is generated by the $D_a$ and $P_{ab}$.
Formally, 
\begin{equation}\label{eqn-maskedphotometricloss}
L_{P}^M = \frac{1}{|\mathcal{V}|} \sum_{p \in \mathcal{V}} (M_s(p) \cdot L_{P} (p)),
\end{equation}
\begin{equation}\label{eqn-photometric2}
L_{P}= \frac{1}{|\mathcal{V}|} \sum_{p \in \mathcal{V}} 
  (\lambda  \Vert I_a(p) - I'_a(p) \Vert _1 + 
  (1-\lambda)  \frac{1-\text{SSIM}_{aa'}(p)}{2} ),
\end{equation}
where $I'_a$ is synthesized from $I_b$ using the warping flow,
and SSIM~\cite{wang2004image} is a widely-used metric to measure image similarity.
We set $\lambda$ to $0.15$ as in \cite{bian2021ijcv}.

Third, we use the edge-aware smoothness loss to regularize the predicted depth map.
Formally,
\begin{equation}\label{eqn:smooth}
L_{S} = \sum_{p} ( e^{-\nabla I_a(p)} \cdot \nabla D_a(p) ) ^2,
\end{equation}
where $\nabla$ is the first derivative along spatial directions, 
which guides smoothness by image edges.

Overall, our objective function is formulated as follows:
\begin{equation}\label{eqn-totalloss}
  L_{Self} = \alpha L_{P}^M + \beta L_{G} + \gamma L_{S}.
\end{equation}
We set $\alpha = 1$, $\beta = 0.5$, and $\gamma = 0.1$ as in \cite{bian2021ijcv}.
Note that we will replace $L_S$ with the proposed normal loss $L_N$ in \secref{sec:structure}.
Moreover, we also use the auto-masking and per-pixel minimum reprojection loss that are proposed in \cite{monodepth2} to filter stationary and non-best points during training.

\begin{figure}[t]
    \centering
    \small
    \begin{tabular}{c c}
     \includegraphics[width=0.45\linewidth]{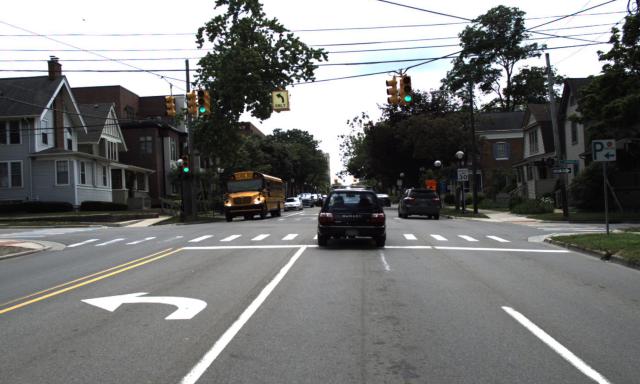}    &  
     \includegraphics[width=0.45\linewidth]{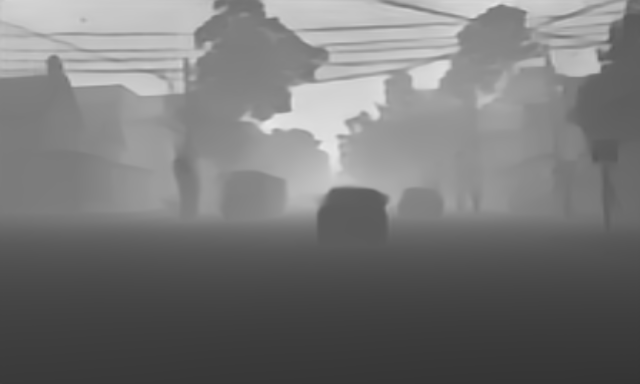} \\
     (a) Image & (b) Pseudo-depth \\
     \multicolumn{2}{c}{
     \includegraphics[width=0.9\linewidth]{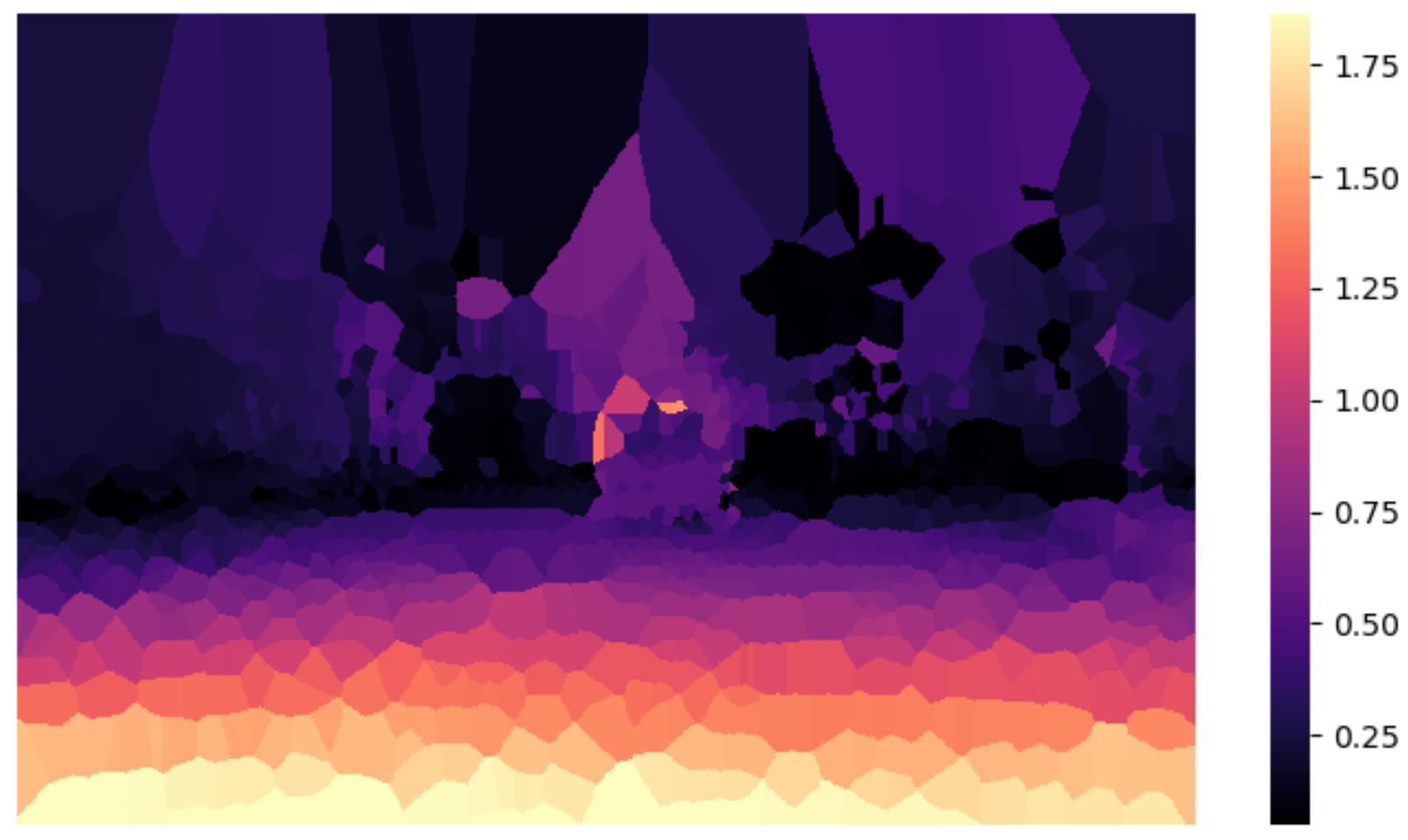}
     }\\
     \multicolumn{2}{c}{(c) Error map (mean AbsRel=0.358)}
     \end{tabular}
    \caption{Visualization of pseudo-depth (LeReS~\cite{yin2021learning}) on the DDAD dataset.
    For the error map, we show the AbsRel error, and we use the nearest interpolation for pixels where the ground-truth depth labels (sparse LiDAR points) are unavailable.
    It shows that LeReS~\cite{yin2021learning} can generalize to previously unseen data with plausible visual results (b),
    however, high quantitative accuracy is not guaranteed. 
    Here ``AbsRel=0.358" is averaged over all testing images. 
    This indicates that our idea of leveraging pseudo-depth for boosting self-supervised training is motivated,
    and it is also non-trivial to use it.
  }
    \label{fig:pseudo-depth-vis}
\end{figure}

\subsection{Single-Image Depth Prior}
\label{sec:prior}

Our idea is to leverage the pretrained monocular depth estimation network for generating single-image depth prior,
which is then used to boost self-supervised learning.
Here we use LeReS~\cite{yin2021learning} to generate \emph{pseudo-depth},
which is trained on large-scale datasets with ground-truth depth labels.
Thanks to the supervised training on large-scale data, 
it shows excellent zero-shot generalization performance on unseen scenes.
Note that LeReS was not trained on datasets that we use in this paper for evaluation.
An example of LeReS outputs is shown in \figref{fig:pseudo-depth-vis},
where it shows plausible visual results on the DDAD dataset but poor quantitative accuracy.

These phenomena echo our motivation, 
\ie, pseudo-depth is not accurate enough but has potential that can be leveraged for boosting self-supervised learning.
More specifically, we find that the good visual results are contributed to several aspects,
including (a) correct depth ordinal (nearer/further relation) between objects;
(b) excellent smoothness in predicted depth;
(c) sharp depth prediction at object boundaries. 

Therefore, based on the above observation,
we propose two modules to extract effective supervision signals from pseudo-depth.
First, we propose a \textit{Dynamic Region Refinement} (DDR) module that regularizes self-supervised training with a depth ranking constraint, 
particularly boosting depth prediction on moving objects.
Second, we propose a \textit{Local Structure Refinement} (LSR) module that constrains the smoothness and object boundaries of the predicted depth.
The proposed two modules are presented in \secref{sec:dynamic} and \secref{sec:structure},
respectively.

\subsection{Dynamic Region Refinement}
\label{sec:dynamic}

\begin{figure}[t]
  \centering
  \includegraphics[width=1.0\linewidth]{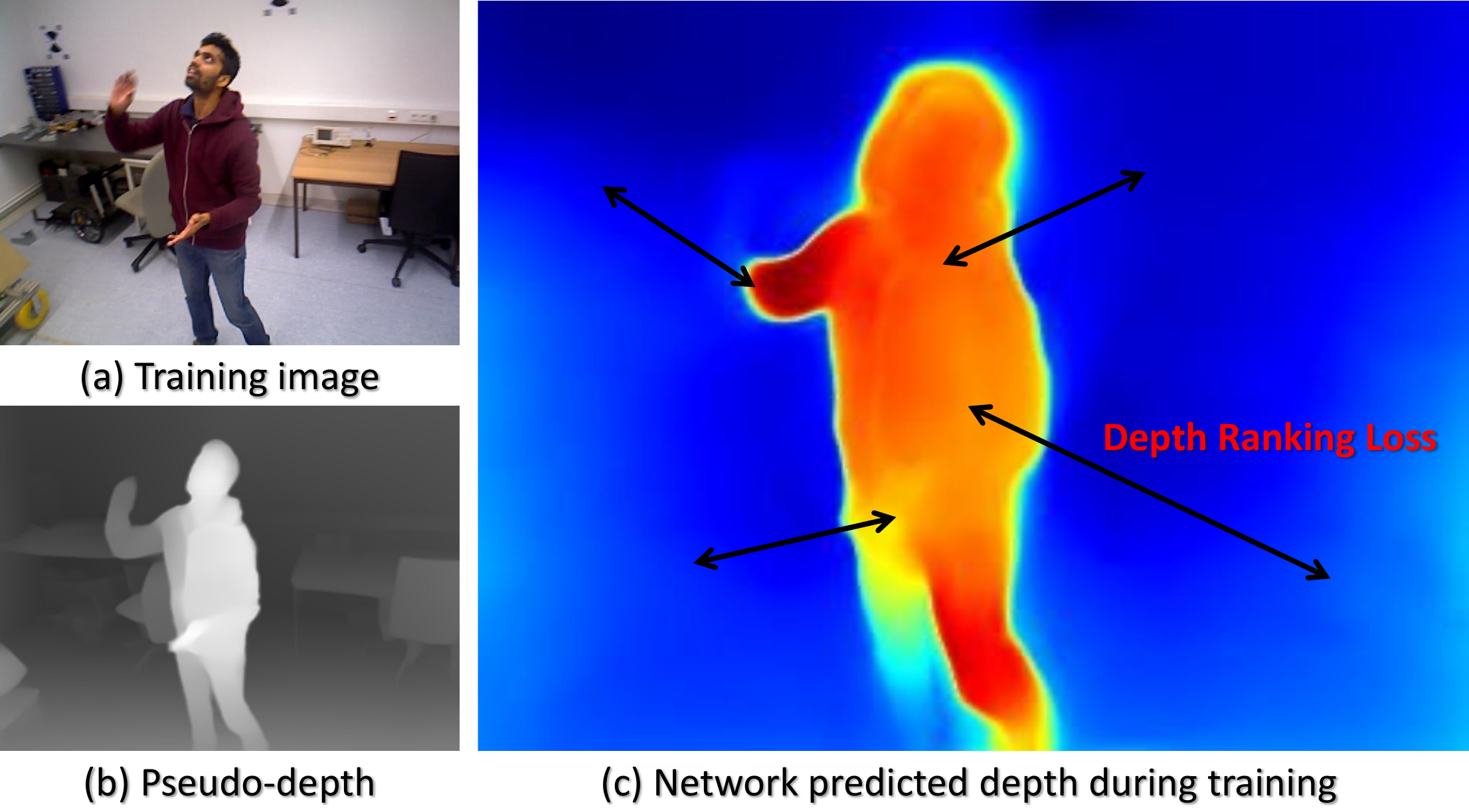}
  \caption{Dynamic region refinement. 
  We sample point pairs between dynamic and static regions, and then we apply depth ranking loss to constrain the network-predicted depth (c) during training.
  The ``ground-truth" depth ordinal is extracted from pseudo-depth (b).
  To segment dynamic regions from static backgrounds, we use the self-discovered mask (\equref{eqn-mask}), so there is no extra computational cost.
  }\label{fig:drr-vis}
\end{figure}

The key to our proposed dynamic region refinement (DRR) module is constraining depth estimation on dynamic regions by enforcing the nearer/further relation \wrt the predicted depths on static regions.
Specifically, this is based on two assumptions including
(i) The accurate depth ranking relations between any two pixels can be extracted from pseudo-depth;
(ii) depth prediction on static regions is sufficiently accurate thanks to the self-supervised losses.
The assumptions are valid,
as demonstrated in prior work~\cite{yin2021learning, monodepth2},
so our idea is generally effective.
In the proposed DRR module,
we first sample point pairs between dynamic and static regions,
where the segmentation of images is obtained in a self-supervised manner (\equref{eqn-mask}).
Then we compute depth ranking loss on sampled point pairs to regularize the predicted depth map.
The proposed sampling method and loss function are presented in the following paragraphs,
and \figref{fig:drr-vis} illustrates a training example.

\paragraph{Dynamic-focused Sampling.}
To sample point pairs between static and dynamic regions,
we need the segmentation of training images.
This could be achieved with the use of pretrained semantic segmentation networks, 
\eg, we can assume objects of certain classes such as vehicles as moving objects and others as static backgrounds.
However, it involves extra data preprocessing and pretrained networks.
Moreover, the dynamic of an object does not necessarily rely on the semantic classes, \eg, a chair can be dynamic while a person is moving it around.
Instead, we derive dynamic/static segmentation from the self-discovered mask (\equref{eqn-mask}) that is computed based on the geometric consistency.
It is a soft weight mask and assigns smaller values for depth-inconsistent regions (dynamics or occlusions) than others (static regions).
To obtain binary segmentation,
we propose to rank weights and pick the lowest $20\%$ as potential dynamic regions,
rather than doing hard thresholding.
Here we assume that the ratio of moving objects pixels is around $20\%$ or less,
which is true in most real-world scenes.
Then for each point in the dynamic regions,
we pair it with a point that is randomly sampled from static regions.
Moreover, other than constructing dynamic-static pairs as discussed above,
we also sample point pairs randomly from the whole image,
which serves as an additional global regularization.

\paragraph{Confident Depth Ranking Loss.}
We compute the depth ranking loss on the sampled point pairs in training.
The original loss function was proposed in~\cite{chen2016single}.
Formally, for a pair of points with predicted depth values
$[p_0, p_1]$, the loss is
\begin{equation}\label{eqn:rank_loss}
    \phi'(p_0, p_1) = 
    \begin{cases}
    \log(1 + \exp(-\ell(p_0 - p_1))),      & \quad \ell \neq 0 \\
    (p_0 - p_1)^2,  & \quad \ell = 0
  \end{cases}
\end{equation}
where $\ell$ is the ground truth ordinal label, which can be induced by a ground truth depth map:
\begin{equation}
    \ell = 
    \begin{cases}
    +1,   & \quad p^*_0 / p^*_1  \geq  1 + \tau, \\
    -1,   & \quad p^*_0 / p^*_1  \leq  \frac{1}{1 + \tau}, \\
    0,    & \quad \rm otherwise. \\
  \end{cases}
\end{equation}
Here $\tau$ is a threshold, which is $0.03$ in previous work~\cite{xian2020structure}, and $p^*$ denotes pseudo-depth.

We empirically find that \equref{eqn:rank_loss} is sub-optimal in our method since pseudo-depth is not as accurate as the ground-truth depth.
Therefore, we have to take the confidence of pseudo-depth ordinals into consideration.
Specifically, we observe that the ordinal is often sufficiently reliable when two points have sufficiently different depth values, \ie, when $p^*_0 / p^*_1 \gg 1 $ or $p^*_0 / p^*_1 \ll 1 $,
and otherwise, it may be unreliable when two depth values are very close,
\ie, when $p^*_0 / p^*_1 \approx 1 $.

Based on the above observation, we propose to (a) increase $\tau$ from $0.03$ to $0.15$ for higher tolerance;
and (b) ignore point pairs that have $\ell = 0$.
Formally, we reformulate \equref{eqn:rank_loss} as
\begin{equation}\label{eqn:new_rank_loss}
    \phi(p_0, p_1) = \log(1 + \exp(-\ell(p_0 - p_1))).
\end{equation}
Therefore, our Confident Depth Ranking Loss is defined as:
\begin{equation}
    L_{\rm CDR} = \frac{1}{|\Omega|} \sum_{p \in \Omega} \phi(p),
\end{equation}
where $\Omega$ stands for the sampled point pairs that have $l \neq 0$.

\subsection{Local Structure Refinement}
\label{sec:structure}

As mentioned above in \secref{sec:prior},
pseudo-depth provides excellent depth smoothness.
In this section, we propose to leverage such attributes to regularize the self-supervised depth.
Our idea is to 
(i) constrain the surface normals that are derived from predicted depths and pseudo-depths to be matched;
and (ii) constrain two depth maps to be consistent \wrt relative normal angles of sampled point pairs around edges.
Here, the first step is focused on refining the overall depth structures, and the second step is focused on improving object boundary regions.
The details are provided below.

\paragraph{Normal Matching Loss.}
The edge-aware depth smoothness loss  (\equref{eqn:smooth}) is often used in self-supervised monocular depth estimation, and it is also used in our baseline.
Here, we propose to replace it with the normal matching loss:
\begin{equation}\label{eqn:normal}
    L_{N} = \frac{1}{N} \sum_{i=1}^{N} ||n_i - n^*_i||_1,
\end{equation}
where $n_i$ is the surface normal derived from the predicted depth,
and $n^*_i$ is the normal derived from pseudo-depth.
$N$ stands for the total number of pixels in the image.
The pixel-wise loss function provides strong supervision for overall depth structures.

\paragraph{Edge-aware Relative Normal Loss.}
Not only do overall structure refinement,
but also we focus on object boundary areas.
Specifically, we sample point pairs around image edges and constrain the relative normal angles of sampled point pairs to be consistent with pseudo-depth.
Here, we use edge-guided sampling that was proposed in~\cite{xian2020structure} to construct point pairs $\langle A$, $B \rangle$,
and we define the Edge-aware Relative Normal Loss as:
\begin{equation}\label{eqn:relative_normal}
    L_{\rm ERN} = \frac{1}{N} \sum_{i=1}^{N} ||n_{Ai} \cdot n_{Bi} - n^*_{Ai} \cdot n^*_{Bi}||_1,
\end{equation}
where $n_{A}$ denotes the normal of a sampled point from the predicted depth,
and $*$ denotes pseudo-depth.
Combining the edge-guided sampling and relative normal loss,
we can effectively constrain the depth estimation on object boundary regions.

The proposed $L_{\rm ERN}$ is similar to the pair-wise normal loss that is proposed in~\cite{yin2021learning},
while the latter samples point pairs from edges, planes, and whole images.
In contrast, we sample points solely from edges because sampling from other regions requires high-quality ground-truth depth.
In our case, pseudo-depth is not accurate enough to maintain high-quality global structures,
hence we only constrain the local structure.
We analyze this effect with a detailed ablation study in \secref{sec:ablation}.

\subsection{Training}

\paragraph{Losses.}
Based on the proposed two refinement modules,
we rewrite the overall loss function (\equref{eqn-totalloss}) of our baseline and obtain the new objective function as:
\begin{equation}\label{eqn-totalloss2}
  L = \alpha L_{P}^M + \beta L_{G} + \gamma L_{N} + \delta L_{\rm CDR} + \epsilon L_{\rm ERN},
\end{equation}
where we set $\alpha = 1$, $\beta = 0.5$, and $\gamma = \delta = \epsilon = 0.1$ in training based on empirical tuning.

\paragraph{Networks.}
Our depth and pose networks are the same as previous work~\cite{bian2021ijcv, monodepth2},
where we use ResNet-18~\cite{he2016deep} backbone for both depth and pose estimation networks.
The depth network is a U-Net structure~\cite{ronneberger2015u} with a DispNet~\cite{zhou2017unsupervised} as the decoder.
The activations are sigmoids at the output layer and ELU nonlinearities~\cite{clevert2015fast} elsewhere. 
We convert the sigmoid output $x$ to depth with $D = 1/(ax + b)$, 
where $a$ and $b$ are chosen to constrain $D$ between 0.1 and 100 units.
The pose network accepts two RGB frames as input and outputs the 6D relative pose.
We modify the first layer of ResNet-18 to have six channels for accepting two-frame inputs,
and features are decoded to 6-DoF parameters via four convolutional layers.

\paragraph{Training Details.}
We implement the proposed method using the PyTorch library~\cite{paszke2017automatic}.
Following~\cite{zhou2017unsupervised,ranjan2019cc,Wang2018CVPR}, 
we use a snippet of three sequential video frames as a training sample.
The images are augmented with random scaling, cropping, and horizontal flips during training.
We use the Adam~\cite{kingma2014adam} optimizer and set the learning rate to be $10^{-4}$.
We initialize the encoder by using the pre-trained model on ImageNet~\cite{imagenet_cvpr09}.
We train our networks in $100k$ iterations on each dataset.


\section{Experiment}
\label{sec:experiment}

\subsection{Datasets and Evaluation Metrics}

The proposed method focuses on boosting self-supervised monocular depth estimation in challenging dynamic scenes,
so we mainly evaluate our methods on three dynamic datasets,
including DDAD driving dataset~\cite{packnet}, BONN dynamic dataset~\cite{palazzolo2019iros}, and TUM dataset~\cite{sturm12iros} (dynamic object split).
Note that these datasets contain fast-moving objects,
which are much more challenging than the widely-used KITTI~\cite{Geiger2013IJRR} and NYUv2~\cite{silberman2012indoor} datasets.
We assume that the latter two datasets are almost static in this paper,
and we also report results on them.
All the mentioned self-supervised methods are trained on each dataset individually for a fair comparison.
Moreover, following previous methods, we analyze the depth results at object boundaries and plane regions in the IBims-1 dataset~\cite{tobias2020cviu}.
In the following paragraphs, the details of each dataset are described.

\begin{table*}[t]
    \centering
    \small
    \caption{Self-supervised monocular depth estimation results on the DDAD driving dataset~\cite{packnet}. We segment vehicles and pedestrians as dynamic objects and consider the remaining regions as static backgrounds. This dataset is more challenging than KITTI due to more complex scenes, fewer stopping cars, and longer depth ranges ($200$m vs $80$m). Note that DynamicDepth~\cite{feng2022disentangling} uses two frames for depth estimation. 
    }\label{tab:ddad_results}
    \vspace{-5pt}
    \begin{tabular}{l | c c c c c c c | c c | c c }
    \toprule
     & \multicolumn{7}{c|}{Full Image} & \multicolumn{2}{c|}{Dynamic} & \multicolumn{2}{c}{Static} \\
     \cline{2-12}
     Methods & AbsRel & SqRel & RMS & RMSlog & $\delta_1$ & $\delta_2$ & $\delta_3$ & AbsRel & $\delta_1$ & AbsRel & $\delta_1$ \\
     \hline

    Monodepth2~\cite{monodepth2}&   0.239  &  12.547   &  18.392  &   0.316  &   0.752  &   0.899  &   0.949 &   0.747  &   0.432 &   0.188 &   0.771 \\
    
     PackNet~\cite{packnet} &   0.182  &   7.945   &  \textbf{15.021}  &   0.259  &   \textbf{0.828}  &   \textbf{0.925}  &   0.961  &   0.564  &  0.520 & \textbf{0.137} &   \textbf{0.843} \\

    SGDepth~\cite{klingner2020self} &   0.200  &   7.944    &  17.149  &   0.289  &   0.769  &   0.911  &   0.957 & 0.619 & 0.446 &0.170 &  0.786 \\
    
    DynamicDepth~\cite{feng2022disentangling} &\textbf{0.156 }& \textbf{3.305} & 15.612 & \textbf{0.258} &0.785 & 0.914 & \textbf{0.962} & \textbf{0.258} & \textbf{0.612} & 0.149 & 0.792 \\
     
    SC-Depth~\cite{bian2021ijcv} &   0.169  &   3.877 &  16.290  &   0.280  &   0.773  &   0.905  &   0.951  & 0.345 & 0.546 & 0.155 & 0.783 \\
    
     \hline
     Ours w/o DRR &   0.153  &   3.124  &  \textbf{15.237}  &   0.252  &   0.799  &   0.920  &   0.963 &   0.259&   0.612 &   0.146&   0.806  \\
     Ours w/o LSR &   0.149  &   3.094  &  16.198  &   0.262  &   0.794  &   0.913  &   0.956 &0.210 &   0.666 &   0.146 &   0.799 \\
     Ours &   \textbf{0.142}  &   \textbf{3.031}  &  15.868  &   \textbf{0.248}  &   \textbf{0.813}  &   \textbf{0.922}  &   \textbf{0.963} & \textbf{0.199} & \textbf{0.697} & \textbf{0.140} & \textbf{0.813} \\
     \bottomrule
  \end{tabular}
  \vspace{-5pt}
\end{table*}

\begin{table*}[t]
    \centering
    \small
    \caption{Self-supervised monocular depth estimation results on the BONN dynamic dataset~\cite{tobias2020cviu}. This dataset is super-challenging because all training and testing videos contain fast-moving objects, which occupy a large proportion of pixels.}\label{tab:bonn_results}
    \vspace{-5pt}
    \begin{tabular}{l | c c  c c c | c c | c c}
     \toprule
     \multirow{2}{*}{Methods} & \multicolumn{5}{c|}{Full Image} & \multicolumn{2}{c|}{Dynamic} & \multicolumn{2}{c}{Static} \\
     \cline{2-10}
      & AbsRel & RMS  & $\delta_1$ & $\delta_2$ & $\delta_3$ & AbsRel & $\delta_1$ & AbsRel & $\delta_1$ \\
     \hline
     Monodepth2~\cite{monodepth2}  &   0.565   &   2.337  &   0.352  &   0.591  &   0.728  &   \textbf{0.474} &   0.172  &   0.594&   0.383 \\
     SC-Depth~\cite{bian2021ijcv}  &   0.272   &   0.733   &   0.623  &   0.858  &   0.948 & 0.704 & 0.166 & 0.180 & 0.714 \\
     SC-DepthV2~\cite{bian2021tpami} &   \textbf{0.211}  &   \textbf{0.619}   &   \textbf{0.714}  &   \textbf{0.873}  &   \textbf{0.936} & 0.488 & \textbf{0.247} & \textbf{0.152} & \textbf{0.803} \\
     \hline
     Ours w/o DRR &   0.138 &   0.396 &   0.885  &   0.951  &   0.974  &   0.248  &   0.690&   0.106 &   \textbf{0.939} \\
     Ours w/o LSR &   0.130  &   0.382   &   0.874  &   0.951  &   0.977  &   0.274 &   0.613 &   0.097  &   0.937 \\
     Ours &   \textbf{0.126} &   \textbf{0.379}  &  \textbf{0.889}  &   \textbf{0.961}  &   \textbf{0.980}  &   \textbf{0.220} &   \textbf{0.720} &   \textbf{0.102} &   0.931 \\
     \bottomrule
    \end{tabular}
    \vspace{-5pt}
\end{table*}

\begin{table*}[t]
    \centering
    \small
    \caption{Self-supervised monocular depth estimation results on the TUM dataset~\cite{sturm12iros}. 
    We use the videos under the category of "Dynamic Objects" for training and testing, in which moving objects occupy a large proportion of pixels in each image.}\label{tab:tum_results}
    \vspace{-5pt}
    \begin{tabular}{l | c c  c c c | c c | c c}
     \toprule
     \multirow{2}{*}{Methods} & \multicolumn{5}{c|}{Full Image} & \multicolumn{2}{c|}{Dynamic} & \multicolumn{2}{c}{Static} \\
     \cline{2-10}
      & AbsRel & RMS  & $\delta_1$ & $\delta_2$ & $\delta_3$ & AbsRel & $\delta_1$ & AbsRel & $\delta_1$ \\
     \hline
     Monodepth2~\cite{monodepth2} &   0.312  &   1.408  &   0.474  &   0.793  &   0.905  &   0.431  &   0.348 &   0.262&   0.526 \\
     SC-Depth~\cite{bian2021ijcv} &   0.257   &   0.283  &   0.616  &   0.814  &  0.909 & 0.512 & 0.274 & \textbf{0.176} & \textbf{0.715} \\
     SC-DepthV2~\cite{bian2021tpami} &   \textbf{0.223}   &   \textbf{0.282}  &   \textbf{0.643}  &   \textbf{0.862}  &   \textbf{0.932} & \textbf{0.283} & \textbf{0.494} & 0.206 & 0.686 \\
     \hline
     Ours w/o DRR  &   0.185 &   1.163  &   0.744  &   \textbf{0.889}  &   \textbf{0.970}  &   0.272  &   0.593 &   \textbf{0.161} &   0.775\\
     Ours w/o LSR &   0.195   &   1.498  &   0.715  &   0.864  &   0.899  &   0.264 &   0.575 &   0.174&   0.759 \\
     Ours  & \textbf{0.163}  & \textbf{0.265}  &   \textbf{0.797}  &   0.882  &   0.937 & \textbf{0.165} & \textbf{0.796} & 0.171 & \textbf{0.780} \\
     \bottomrule
    \end{tabular}
    \vspace{-5pt}
\end{table*}

\paragraph{DDAD.}
The dataset contains $200$ driving videos that are captured in urban scenes.
The LiDAR scanned point clouds are provided,
which we use to generate sparse ground-truth depths for evaluation.
In this dataset, almost vehicles are moving on the road,
and there are fewer stopping cars than KITTI,
making it more challenging to train self-supervised models.
We use the standard training/testing split, which has 150 training scenes (12650 images) and 50 validation scenes (3950 images).
We use the validation scenes for evaluation.
Depth ranges are capped to at most $200$ meters,
and images are resized to the resolution of $640 \times 384$ for training depth and pose networks.

\begin{figure*}[t]
    \centering
    \small
    \begin{tabular}{c c}
     \includegraphics[width=0.48\linewidth]{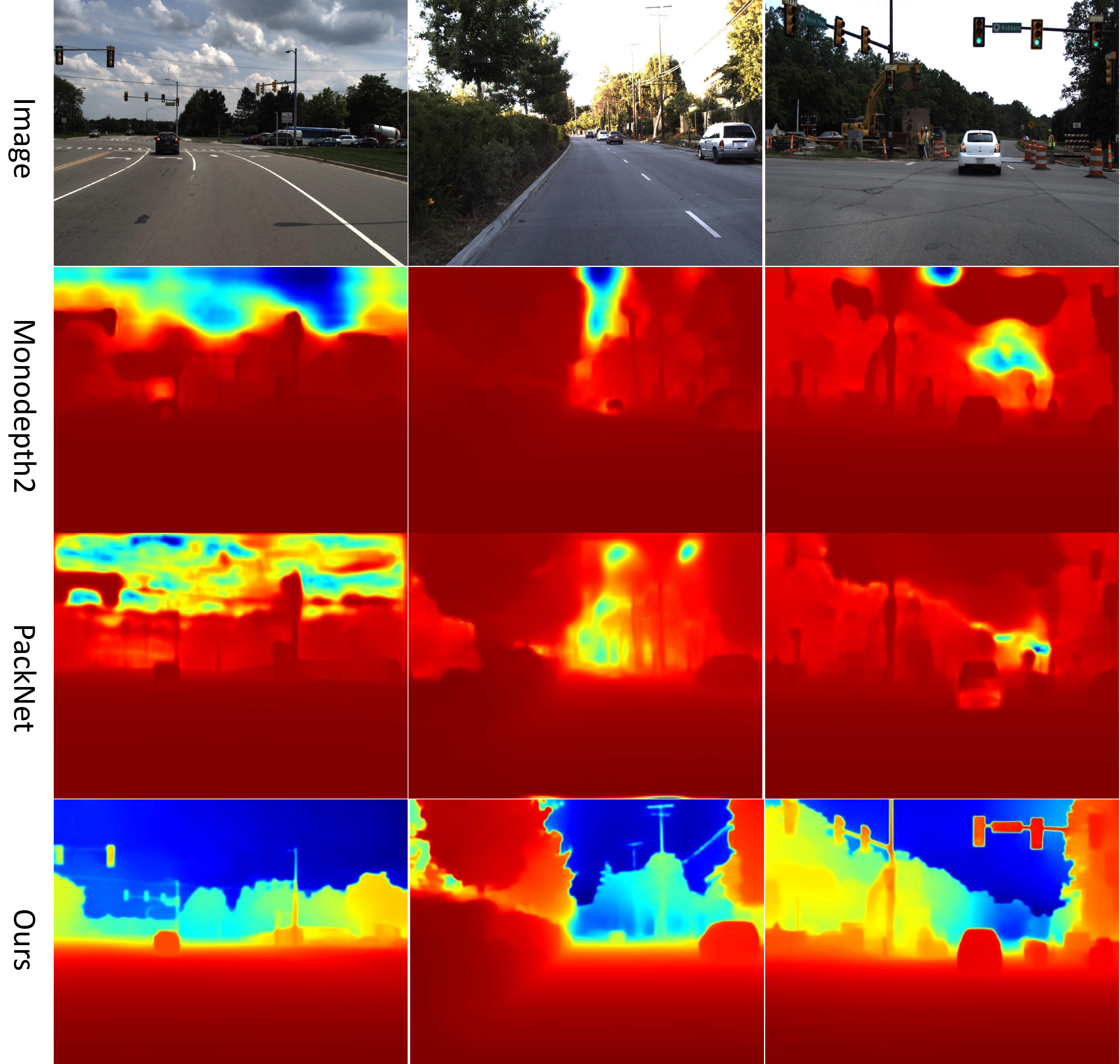}    &  
     \includegraphics[width=0.48\linewidth]{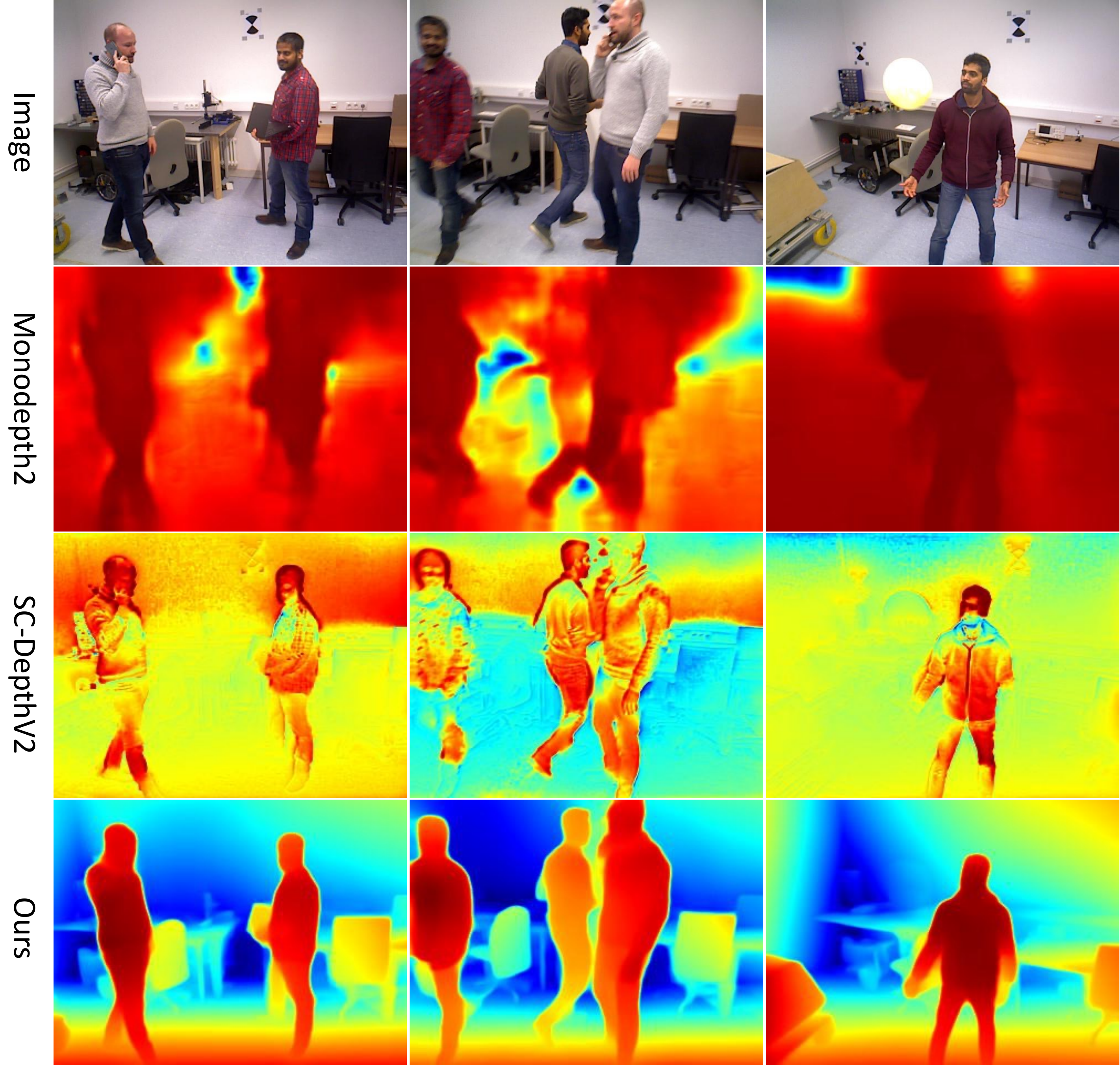} \\
     (a) DDAD & (b) BONN \\
     \includegraphics[width=0.48\linewidth]{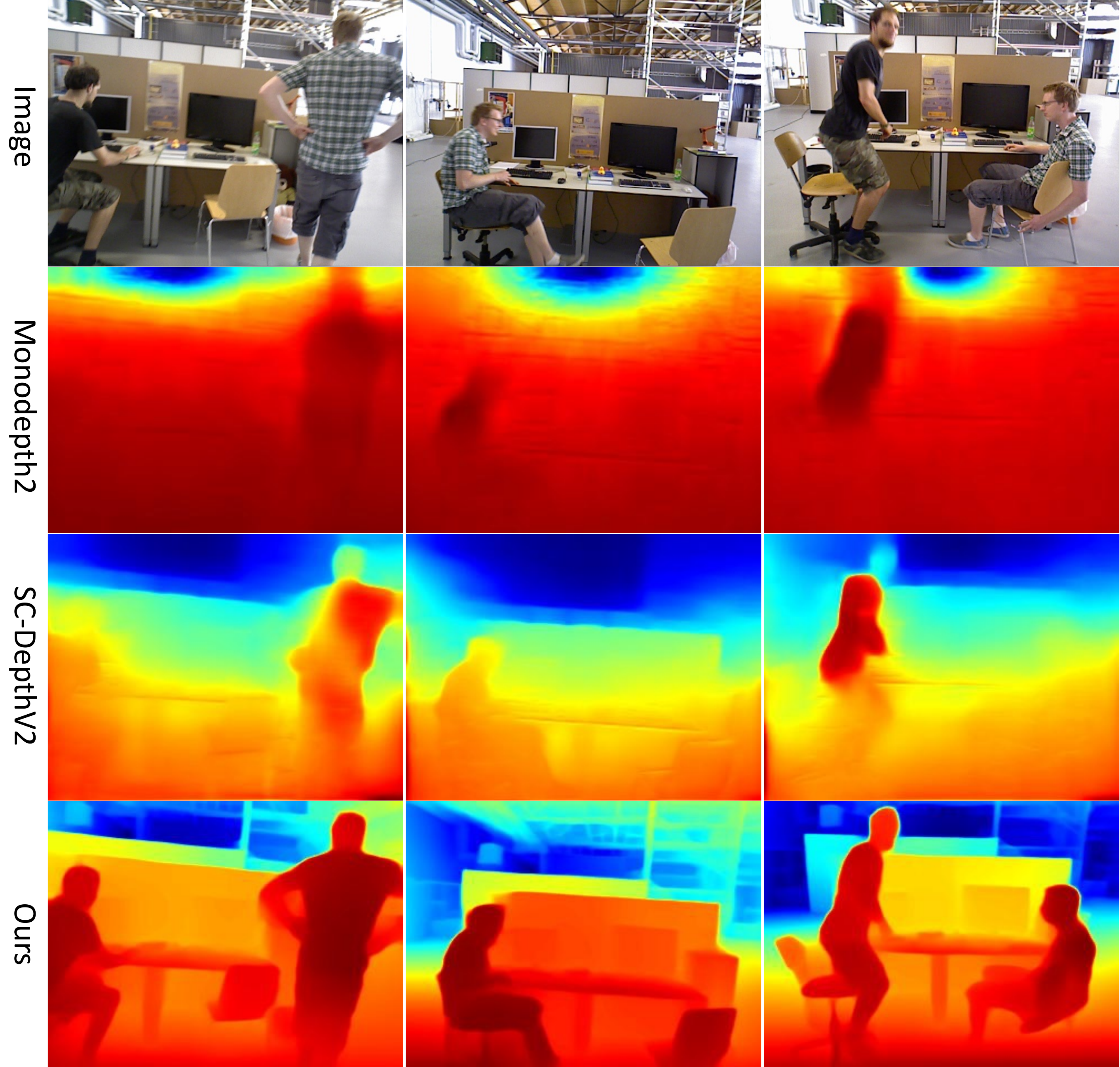}    &  
     \includegraphics[width=0.48\linewidth]{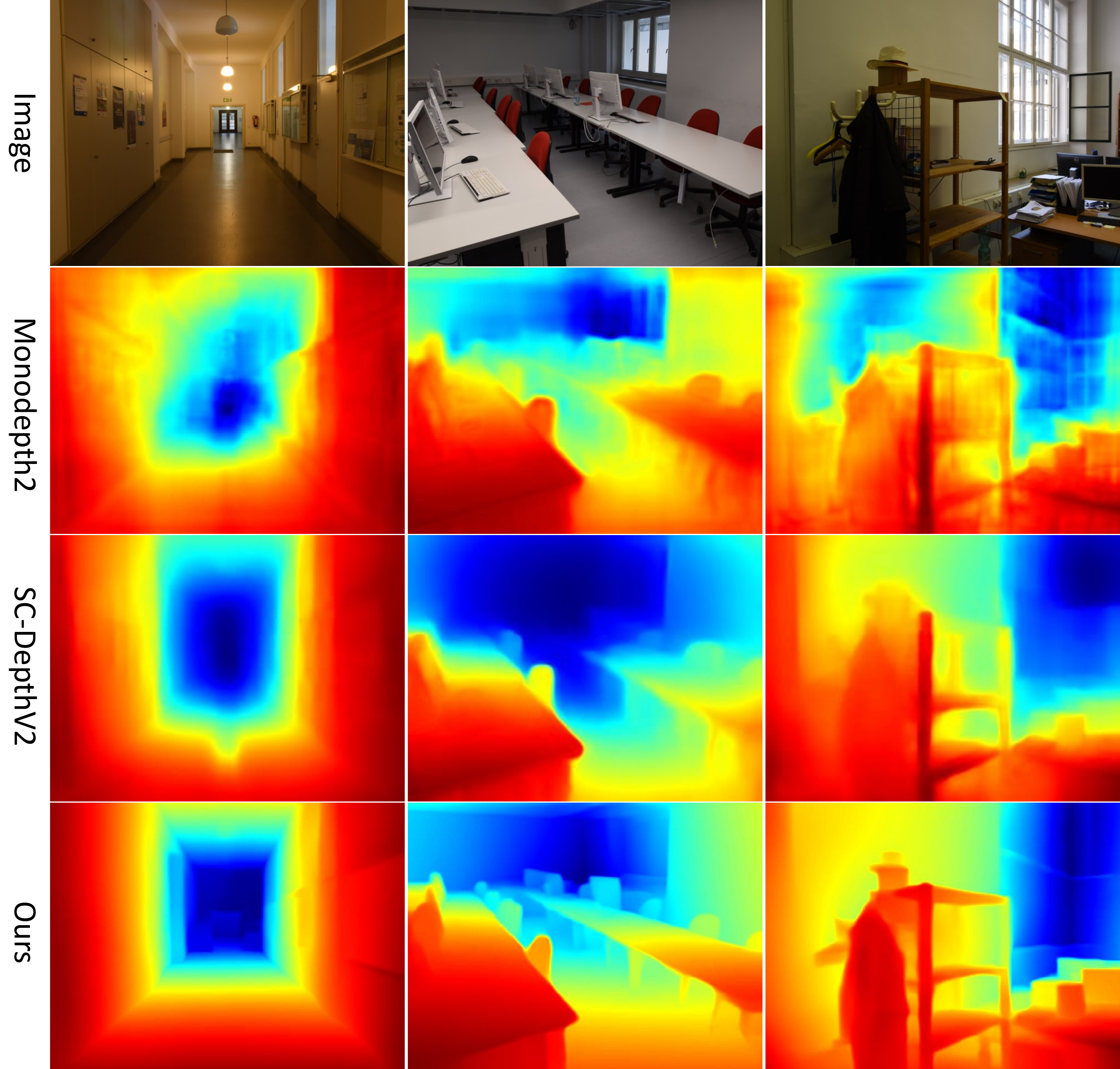} \\
     (c) TUM & (d) IBims-1 (Models trained on NYUv2) \\
    \end{tabular}
    \caption{Qualitative depth estimation results. Existing methods show poor results in dynamic scenes (a-c) because they are hard to handle fast-moving objects during training. Even though they show good accuracy in (d), where models are trained in static scenes, the depth is blurred at object boundaries. By contrast, our method predicts sharp and accurate depth robustly.  
    }
    \label{fig:vis_compare}
\end{figure*}

\paragraph{BONN.}
The dataset contains $26$ dynamic indoor videos that have fast-moving people or other objects.
The Kinnect captured depth maps are provided as the ground truth for evaluation.
We manually find $4$ challenging video sequences with fast-moving people ($1785$ images) for testing,
and we use the remaining videos for training.
Depth ranges are capped at $10$ meters,
and images are resized to the resolution of $320 \times 256$ for training networks.

\paragraph{TUM.}
The dataset provides a collection of indoor videos with Kinnect-captured depth maps as the ground truth
We choose only videos that belong to the \emph{Dynamic Objects} category,
making sure that the model is trained in dynamic scenes.
There are in total $11$ sequences, 
and we use the last two sequences that contain moving people ($1375$ images) for testing.
The remaining $9$ dynamic videos are used for training,
and images are resized to the resolution of $320 \times 256$ for feeding to networks.

\paragraph{KITTI.}
The dataset provides driving videos in urban scenes,
and it is the most widely-used dataset in self-supervised monocular depth estimation problems.
Following previous work~\cite{zhou2017unsupervised, monodepth2, bian2021ijcv, packnet},
we use the Eigen's split that has $697$ images for testing,
and we use the remaining video sequences for training.
Depth ranges are capped at $80$ meters,
and images are resized to the resolution of $832 \times 256$ for training networks.
Note that KITTI contains a large number of stopping cars that help self-supervised methods learn depth estimation on cars,
so the results on this dataset cannot reflect our main contributions,
\ie, robust learning of monocular depth from dynamic scenes.

\paragraph{NYUv2.}
The dataset provides a large collection of indoor videos,
and it is widely-used in the computer vision community.
There are $654$ testing images of static scenes for depth evaluation,
and we use the remaining videos that do not contain testing images for training neural networks.
Images are resized to the resolution of $320 \times 256$ before feeding to the network.
Note that this dataset contains almost-static scenes.

\begin{table*}[t]
  \centering
  \small
  \caption{Self-supervised monocular depth estimation results on KITTI~\cite{packnet}. Note that the KITTI dataset has many stopping vehicles that help learn depth on cars, which is not the case of \emph{learning dynamic object depth from dynamic video} that we addressed in this paper. Besides, note that PackNet uses a large backbone, while other methods including ours use the ResNet-18 encoder.
    }\label{tab:kitti_results}
  \begin{tabular}{l | c c c c c c c | c c | c c }
    \toprule
     & \multicolumn{7}{c|}{Full Image} & \multicolumn{2}{c|}{Dynamic} & \multicolumn{2}{c}{Static} \\
     \cline{2-12}
     Methods & AbsRel & SqRel & RMS & RMSlog & $\delta_1$ & $\delta_2$ & $\delta_3$ & AbsRel & $\delta_1$ & AbsRel & $\delta_1$ \\
     \hline

     Monodepth2~\cite{monodepth2} &   0.114  &   0.848  &   4.986  &   0.198  &   0.869  &   0.956  &   0.980   &   \textbf{0.187} &   0.731 &   0.104   &   0.884\\
    
     PackNet~\cite{packnet} &   \textbf{0.109}  &   \textbf{0.839}  &   \textbf{4.696}  &   \textbf{0.188}  &   \textbf{0.884}  &   0.961  &   0.981 &   0.208  &   \textbf{0.737} &   \textbf{0.099} &   \textbf{0.901} \\

    SGDepth~\cite{klingner2020self}   &   0.111  &   0.857   &   4.739  &   0.189  &   \textbf{0.884}  &   \textbf{0.962}  &   \textbf{0.982}  & 0.209 &  0.728 & 0.101 & 0.899 \\
     
     SC-Depth~\cite{bian2021ijcv} &   0.118  &   0.870  &   4.997  &   0.196  &   0.860  &   0.956  &   0.981 &   0.242&   0.698&   0.108 &   0.878\\
     \hline
     Ours &   \textbf{0.118}  &   \textbf{0.756}  &   \textbf{4.709}  &   \textbf{0.188}  &   \textbf{0.864}  &   \textbf{0.960}  &   \textbf{0.984}  & \textbf{0.205} & \textbf{0.703} & \textbf{0.108} & \textbf{0.881} \\
     \bottomrule
  \end{tabular}
\end{table*}

\begin{table}[t]
    \centering
    \footnotesize
    \caption{Monocular depth estimation results on the NYUv2~\cite{silberman2012indoor} dataset. 
    Our method outperforms a majority of supervised methods (first row) and all the self-supervised methods (second row).}
    \label{tab:nyu_results}
    \begin{tabular}{l | c c  | c c c}
     \toprule[1pt]
     \multirow{2}{*}{Methods} & \multicolumn{2}{c|}{Error $\downarrow$} & \multicolumn{3}{c}{Accuracy $\uparrow$}  \\
     \cline{2-6}
      & AbsRel & RMS  & $\delta_1$ & $\delta_2$ & $\delta_3$ \\
     \hline
     Make3D~\cite{saxena2006learning} &  0.349 & 1.214 & 0.447 & 0.745 & 0.897 \\
     DepthTransfer~\cite{karsch2014depth}  & 0.349 & 1.210 & - & - & - \\
     Liu~\etal~\cite{liu2014discrete} &  0.335 & 1.060 & - & - & - \\ 
     Ladicky~\etal~\cite{ladicky2014pulling} & - & - & 0.542 & 0.829 & 0.941 \\
     Li~\etal~\cite{li2015depth} & 0.232 & 0.821 & 0.621 & 0.886 & 0.968 \\
     Roy~\etal~\cite{roy2016monocular} &  0.187 & 0.744 & - & - & - \\
     Wang~\etal~\cite{wang2015towards}  & 0.220 & 0.745 & 0.605 & 0.890 & 0.970 \\
     Eigen~\etal~\cite{eigen2015predicting} & 0.158 & 0.641 & 0.769 & 0.950 & 0.988 \\
     Chakrabarti~\etal~\cite{chakrabarti2016depth} & 0.149 & 0.620 & 0.806 & 0.958 & 0.987 \\
     Li~\etal~\cite{li2017two} & 0.143 & 0.635 & 0.788 & 0.958 & 0.991 \\
     DORN~\cite{fu2018deep} & 0.115 & 0.509 & 0.828 & 0.965 & 0.992 \\
     VNL~\cite{Yin2019enforcing} & \textbf{0.108} & \textbf{0.416} & \textbf{0.875} & \textbf{0.976} & \textbf{0.994} \\
     \hline
     Zhou~\etal~\cite{Zhou_2019_ICCV}  & 0.208 & 0.712 & 0.674 & 0.900 & 0.968 \\
     Zhao~\etal~\cite{zhao2020towards}  & 0.189 & 0.686 & 0.701 & 0.912 & 0.978 \\
    Monodepth2~\cite{monodepth2} & 0.169 &   0.614   &   0.745  &   0.946  &   0.987  \\
     SC-Depth~\cite{bian2021ijcv}  & 0.159 & 0.608 & 0.772 & 0.939 & 0.982 \\
     P2Net~\cite{IndoorSfMLearner}  & 0.150 & 0.561 & 0.796 & 0.948 & 0.986 \\
    SC-DepthV2~\cite{bian2021tpami} & 0.138 & 0.532 & 0.820 & 0.956 & \textbf{0.989}\\
     MonoIndoor~\cite{ji2021monoindoor}  & \textbf{0.134} & \textbf{0.526} & \textbf{0.823} & \textbf{0.958} & \textbf{0.989} \\
     \hline
     Ours  &   \textbf{0.123}  &   \textbf{0.486}  &   \textbf{0.848}  &   \textbf{0.963}  &   \textbf{0.991} \\
     \bottomrule
    \end{tabular}
\end{table}

\begin{table}[t]
\centering
\caption{Evaluation of depth boundaries (DBE) and planes (PE) on iBims-1~\cite{tobias2020cviu}. All models are trained on NYUv2.}
\label{tab:ibims}

    \begin{tabular}{l|ccccc}
    \toprule[1pt]
    \multirow{2}{*}{Method} & \multicolumn{4}{c}{iBims-1} \\ 
    \cline{2-6} 
    & $\varepsilon^\text{acc}_\text{DBE}\downarrow$ & \multicolumn{1}{l|}{$\varepsilon^\text{comp}_\text{DBE}\downarrow$} & $\varepsilon^\text{plan}_\text{PE}\downarrow$ & \multicolumn{1}{l|}{$\varepsilon^\text{orie}_\text{PE}\downarrow$} &AbsRel$\downarrow$\\ 
    \hline
    Monodepth2~\cite{monodepth2} 
& 4.269
& 89.771
& 10.943
& 29.327
& 0.202\\

SC-DepthV2~\cite{bian2021tpami} 
& \textbf{4.206}
& \textbf{69.846}
& \textbf{7.049}
& \textbf{23.109}
& \textbf{0.172}
\\

\hline
Ours w/o LSR
& 3.138
& 65.692
& 3.684
& 14.696
& 0.152
\\
Ours 
& \textbf{3.001}
& \textbf{48.047}
& \textbf{2.701}
& \textbf{13.372}
& \textbf{0.146}
\\
\bottomrule
\end{tabular}
\end{table}

\paragraph{IBims-1.}
The dataset provides $100$ accurate and dense ground truths for analyzing depth details,
including object boundaries and planes.
Images are collected in different kinds of indoor environments,
and it does not provide a training set.
For a fair comparison with previous work, we use the model trained on the NYUv2 dataset for all methods.

\paragraph{Depth Evaluation Metrics.}
We use standard depth evaluation metrics, including mean absolute relative error (AbsRel), root mean squared error (RMS), root mean squared log error (RMSlog), and the accuracy under threshold ($\delta_i$ $<$ $1.25^i$, $i = 1, 2, 3$).
The detailed definition of these depth metrics can be found in \cite{eigen2014depth}.
Besides, following previous work~\cite{zhou2017unsupervised, bian2021ijcv},
we multiply the predicted depth maps by a scalar that matches the median with that of the ground truth for evaluation,
\ie, $s = median(D_{gt})/median(D_{pred})$,
since self-supervised methods cannot recover the metric scale. For the evaluation on iBims, the depth boundary errors (DBE) and planarity errors (PE) are used to evaluate the accuracy of depth boundaries and planarity respectively. The detailed definitions of DBE and PE are in ~\cite{tobias2020cviu}.

\paragraph{Evaluation on Static/Dynamic Regions.}
We use MSeg~\cite{lambert2020mseg} to generate the semantic segmentation mask of testing images.
The model is trained on a composite dataset, so it is able to generate segmentation results for both indoor and outdoor driving scenes. 
In driving datasets (\ie, KITTI and DDAD), all vehicle and pedestrian segments are regarded as dynamic objects,
and other regions are regarded as static backgrounds.
In indoor datasets (\ie, TUM and BONN), we consider all human segments as dynamic regions.
Note that we align the global scale to the ground-truth depth first,
and then we evaluate depth accuracy on static regions, dynamic regions, and full images, individually.

\subsection{Evaluation Results}

\paragraph{Results on Dynamic Datasets.}
We use three dynamic datasets mentioned above to evaluate the proposed method,
and the quantitative depth estimation results are reported in \tabref{tab:ddad_results},
\ref{tab:bonn_results}, and \ref{tab:tum_results}, respectively.
We show the qualitative comparison results in \figref{fig:vis_compare},
and demo videos for depth estimation are in the supplementary.
A more detailed analysis is conducted below.

\tabref{tab:ddad_results} shows the results on DDAD dataset,
where we compare our method with previous state-of-the-art methods,
including Monodepth2~\cite{monodepth2}, PackNet~\cite{packnet}, and SC-Depth~\cite{bian2021ijcv}.
The results show that our method outperforms previous methods by a large margin,
and particularly on dynamic regions.
Note that our method outperforms PackNet~\cite{packnet},
although the latter uses a significantly larger network backbone than ours.
This demonstrates our main contribution in this paper, \ie, robust learning of monocular depth in dynamic scenes.
Besides, we also report the result without our proposed DRR and LSR modules.
Here our baseline method is a modified version of SC-Depth,
and it incorporates the advantages of Monodepth2.
The results show that the performance of these models is significantly lower than that of our full model,
which demonstrates the efficacy of our proposed losses.

\tabref{tab:bonn_results} and \tabref{tab:tum_results} show the depth estimation results on BONN and TUM datasets, respectively.
These indoor datasets are more challenging than driving datasets such as DDAD
since the ratio of dynamic regions to the full image of the former is significantly larger than that of the latter.
Consequently, previous methods such as Monodepth2~\cite{monodepth2} and SC-Depth~\cite{bian2021ijcv} show poor accuracy in BONN and TUM datasets.
Compared with these approaches, our method presents significantly better results.
This is contributed to our proposed losses,
which enables our method to learn depth estimation robustly from dynamic videos.

\begin{table}[t]
  \centering
  \caption{Ablation studies of the proposed DRR on DDAD dataset. RS denotes random sampling used in ~\cite{chen2016single}, and RL denotes ranking loss used in ~\cite{chen2016single}. The decreased performance demonstrates the effectiveness of our proposed methods.
  }\label{tab:ddad_drr}
  \resizebox{0.99\linewidth}{!}{%
  \begin{tabular}{l | c c | c c | c c }
     \toprule[1pt]
     & \multicolumn{2}{c|}{Full} & \multicolumn{2}{c|}{Dynamic} & \multicolumn{2}{c}{Static} \\
     \cline{2-7}
     Methods & AbsRel & $\delta_1$ & AbsRel & $\delta_1$ & AbsRel & $\delta_1$ \\
     \hline
     Baseline &   0.179 & 0.753 & 0.355 & 0.536 & 0.163 & 0.761 \\
     B+DRR (Ours) &  \textbf{0.149}  & \textbf{0.794} & \textbf{0.210} &   \textbf{0.666} &   \textbf{0.146} & \textbf{0.799} \\
     DRR w/ RS &   0.154 & 0.785 & 0.219 &   0.654 &   0.151&   0.790  \\
     DRR w/ RL &   0.159 & 0.767 & 0.214 &   0.659 &   0.159 &   0.765 \\
     \bottomrule
  \end{tabular}
  }
\end{table}

\begin{table}[t]
  \centering
  \caption{Ablation studies of the proposed LSR on DDAD dataset. EDS denotes edge-aware depth smoothness~\cite{bian2021ijcv}, and RS denotes additional random sampling beside edge-based sampling. The decreased performance demonstrates the importance of our proposed methods.
  }\label{tab:ddad_lsr}
  \resizebox{0.99\linewidth}{!}{%
  \begin{tabular}{l | c c | c c | c c }
     \toprule[1pt]
     & \multicolumn{2}{c|}{Full} & \multicolumn{2}{c|}{Dynamic} & \multicolumn{2}{c}{Static} \\
     \cline{2-7}
     Methods & AbsRel & $\delta_1$ & AbsRel & $\delta_1$ & AbsRel & $\delta_1$ \\
     \hline
     Baseline &   0.179 & 0.753 & 0.355 & 0.536 & 0.163 & 0.761 \\
     LSR (Ours)  &  \textbf{0.142}  & \textbf{0.813} & \textbf{0.199} &   \textbf{0.697} &   \textbf{0.140} & \textbf{0.813} \\
     LSR w/ EDS &   0.148 & 0.793 & 0.200 &   0.694 &   0.145&   0.796  \\
     LSR w/ RS &0.146 & 0.802 & 0.200 & 0.688 & 0.143 & 0.806 \\
     \bottomrule
  \end{tabular}
  }
\end{table}

\begin{figure*}[t]
    \centering
    \small
    \begin{tabular}{c c}
     \includegraphics[width=0.48\linewidth]{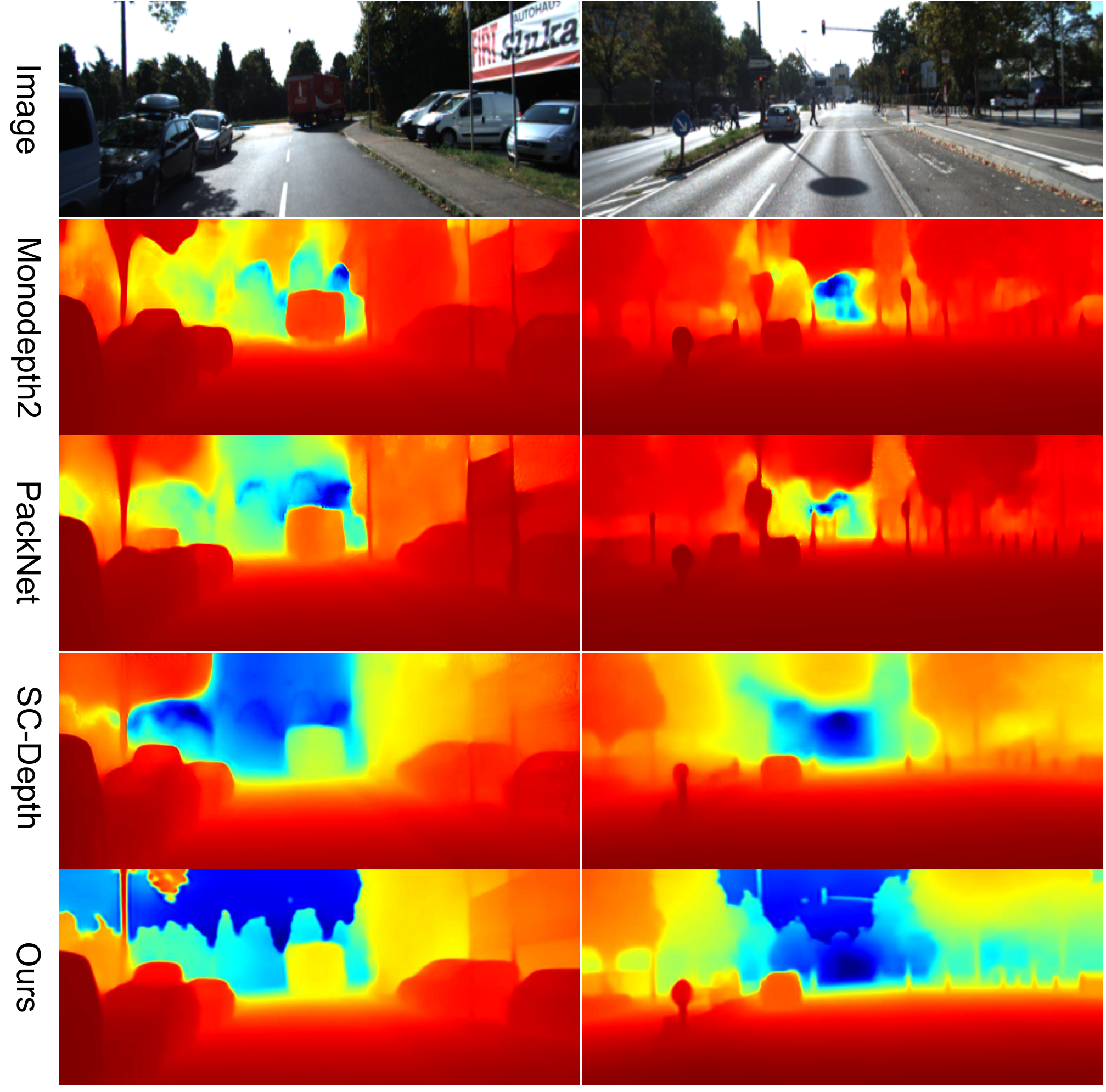}    &  
     \includegraphics[width=0.48\linewidth]{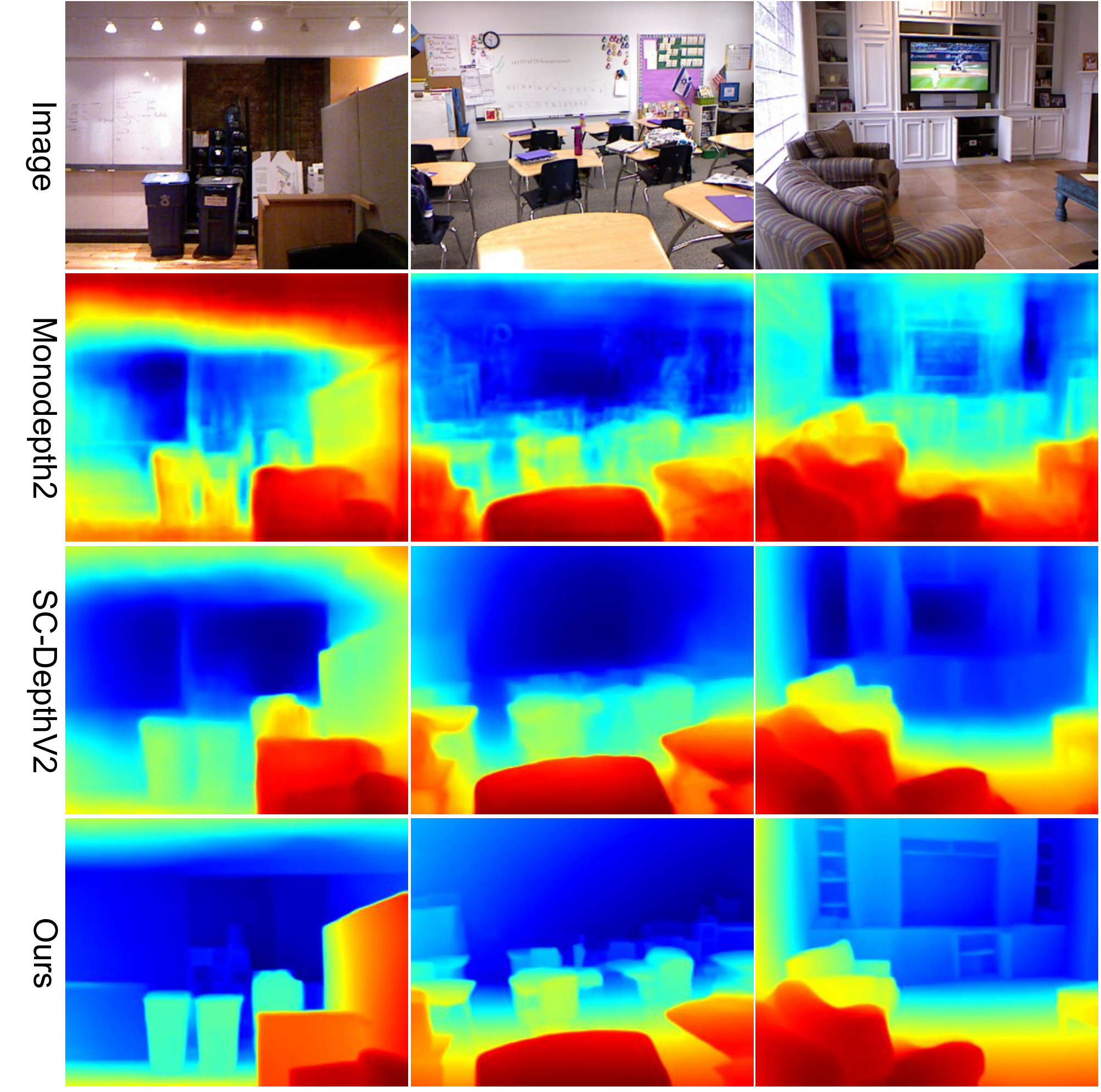} \\
     (a) KITTI & (b) NYUv2
    \end{tabular}
    \caption{Qualitative monocular depth estimation results on static datasets. Our method allows for generating sharper depth maps than previous methods---See object boundaries.
    }
    \label{fig:vis_static}
\end{figure*}

\paragraph{Results on static Datasets.}
Although our main contribution in this paper is boosting self-supervised monocular depth in dynamic scenes,
we show that our method is also working well in almost-static scenes.
The results are reported in the widely-used KITTI driving dataset and NYUv2 indoor dataset.
Sampled qualitative results are illustrated in \figref{fig:vis_static}.

\tabref{tab:kitti_results} shows the depth estimation results on KITTI,
where our method is comparable but does not outperform the previous state-of-the-art methods.
The reasons are two folds.
First, the dataset contains a large number of stopping cars that help self-supervised methods learn depth on vehicles,
so our method is hard to further improve the performance when previous methods have obtained good results on dynamic regions.
Second, PackNet~\cite{packnet} uses a large network backbone, 
while other methods, including ours, use ResNet-18,
which is much smaller than the former.
Overall, we argue that the existing methods have reached a bottleneck in the KITTI dataset,
and due to the low impact of dynamic objects on self-supervised learning here,
our method is hard to further improve the performance.
Moreover, we show qualitative results in \figref{fig:vis_static} (a),
which shows that our method generates sharper depth maps than other methods.

\tabref{tab:nyu_results} shows the depth results on NYUv2,
where we compare our method with previous state-of-the-art methods such as SC-DepthV2~\cite{bian2021tpami} and MonoIndoor~\cite{ji2021monoindoor}.
The results show that our method outperforms previous approaches significantly.
This is mainly contributed to the single-image depth prior,
which we use to constrain the normal smoothness and sharp object boundaries of predicted depths.
The qualitative results are shown in \figref{fig:vis_static} (b) and \figref{fig:vis_compare} (d),
and the quantitative evaluation results on object boundaries are summarized in \tabref{tab:ibims},

\paragraph{Depth Quality at Object Boundaries.}
\tabref{tab:ibims} shows the detailed analysis of depth results on the IBims-1 dataset,
where we compare our method with Monodepth2~\cite{monodepth2} and SC-DepthV2~\cite{bian2021tpami}.
All models are trained on NYUv2 for a fair comparison.
The AbsRel metric shows the overall accuracy of depth estimation results on full images,
and other metrics reflect the detailed depth quality at object boundaries and plane regions.
The results show that our method significantly outperforms previous methods.
We also remove the proposed LSR from our full model for ablation study purposes,
and the results in \tabref{tab:ibims} show that the performance is clearly degraded.
This demonstrates the efficacy of our proposed LSR module.
The qualitative depth estimation results on the IBims-1 dataset are illustrated in \figref{fig:vis_compare} (d).

\begin{table}[t]
  \centering
  \caption{Evaluation results on DDAD dataset. We compare different methods for generating pseudo-depth. ``+Self'' means training models with our proposed self-supervised method.
  }\label{tab:ddad_pseudo}
  \resizebox{0.99\linewidth}{!}{%
  \begin{tabular}{l | c c | c c | c c }
     \toprule[1pt]
     & \multicolumn{2}{c|}{Full Image} & \multicolumn{2}{c|}{Dynamic} & \multicolumn{2}{c}{Static} \\
     \cline{2-7}
     Methods & AbsRel & $\delta_1$ & AbsRel & $\delta_1$ & AbsRel & $\delta_1$ \\
     \hline
     DPT~\cite{Ranftl2021} & 0.224 & 0.632 & 0.296 & 0.492 & 0.220 & 0.636 \\
     DPT+Self & \textbf{0.151} & \textbf{0.788} & \textbf{0.218} & \textbf{0.662} & \textbf{0.147} & \textbf{0.791} \\
     \hline
     LeReS(Res50)~\cite{yin2021learning} & 0.385 & 0.411 & 0.354 & 0.380 & 0.390 & 0.402 \\
     LeReS(Res50)+Self & \textbf{0.147} & \textbf{0.797} & \textbf{0.188} & \textbf{0.726} & \textbf{0.145} & \textbf{0.798} \\
     \hline
     LeReS(Res101)~\cite{yin2021learning} &  0.358 & 0.434  & 0.341 & 0.386 & 0.363 & 0.424\\
     LeReS(Res101)+Self (Ours)&  \textbf{0.142}  &    \textbf{0.813}  & \textbf{0.199} & \textbf{0.697} & \textbf{0.140} & \textbf{0.813} \\
     \bottomrule
  \end{tabular}
  }
\end{table}

\subsection{Ablation Studies} \label{sec:ablation}

We have shown results with and without our proposed DRR and LSR in \tabref{tab:ddad_results}, \ref{tab:bonn_results} and \ref{tab:tum_results}.
The results demonstrate the efficacy of the proposed modules.
In this section, we make a more detailed analysis of the proposed methods,
and we also discuss the performance by using different methods to generate pseudo-depth.

\paragraph{Dynamic Region Refinement.}
The proposed DRR module consists of dynamic-focused sampling and confident depth ranking loss.
We make ablation studies by comparing our method with random sampling (RS) and original ranking loss (RL) that are used in \cite{chen2016single}.
\tabref{tab:ddad_drr} shows the evaluation results,
which show that the performance is significantly degraded when replacing our proposed terms with the existing methods.
This demonstrates the efficacy of our proposed methods.

\paragraph{Local Structure Refinement.}
The proposed LSR module consists of normal matching loss and edge-guided relative normal ranking loss.
The ablation study results are summarized in \tabref{tab:ddad_lsr}.
We replace the normal matching loss with edge-aware depth smoothness loss (EDS),
and we also add random sampling (RS) to the edge-guided sampling.
These variants degenerate the depth accuracy,
which demonstrates that our proposed methods are better than existing solutions.

\paragraph{Pseudo-depth.}
We use LeReS~\cite{yin2021learning} (ResNet-101) in this paper for generating pseudo-depth,
while it is also possible to use other monocular depth estimation networks.
\tabref{tab:ddad_pseudo} shows the ablation study results on DDAD dataset,
where we also include DPT~\cite{Ranftl2021} and ResNet-50 version of LeReS.
The results show that the pseudo-depths that are generated by all three variants are not accurate in the DDAD dataset.
However, when applying our proposed method that uses pseudo-depth for training self-supervised models,
high-accuracy depth estimation results can be obtained.
This demonstrates that our proposed method is not limited to one specific method for generating pseudo-depth.
The results also show that our method with LeReS (ResNet-101)~\cite{yin2021learning} outperforms other variants, including DPT.
We hypothesize that the reason is that our method incorporates the normal information in training which is shared with LeReS but not with DPT.

\paragraph{Discussion.}
We use pseudo-depth to boost self-supervised monocular depth estimation,
which somewhat degrades our claim of self-supervised learning.
However, in practice, the monocular depth estimation models such as~\cite{yin2021learning, Ranftl2021} are only trained once in large-scale datasets and can be used as off-the-shelf tools in new unseen scenes,
so our method has almost no extra cost compared to pure self-supervised depth estimation methods~\cite{bian2021ijcv, monodepth2}.


\section{Conclusion}

We propose SC-DepthV3 for robust self-supervised learning of monocular depth from challenging dynamic videos.
The key to our method is that we use pseudo-depth, which is generated by a pretrained monocular depth estimation network, 
for addressing the challenges in self-supervised monocular depth estimation framework.
More specifically, we address the issues of dynamic objects and blurred object boundaries. 
As a result, our proposed method can predict sharp and accurate depth maps,
even when the model is trained from highly dynamic videos.
We comprehensively evaluate our method on six challenging datasets,
including both dynamic and static scenes.
The results show that our method significantly outperforms previous alternatives,
and the ablation study results demonstrate that the proposed modules are effective.

\ifCLASSOPTIONcompsoc
  \section*{Acknowledgments}
\else
  \section*{Acknowledgment}
\fi
This work was in part supported by National Key R\&D
Program of China (No.\ 2022ZD0118700).
This work was in part supported by the Australian Centre of Excellence for Robotic Vision CE140100016, 
and the ARC Laureate Fellowship FL130100102 to I. Reid. 
We thank anonymous reviewers for their valuable suggestions.

\ifCLASSOPTIONcaptionsoff
  \newpage
\fi

\bibliographystyle{IEEEtran}
\bibliography{reference}

\end{document}